\documentclass[11pt,a4paper]{article}
\usepackage[hyperref]{acl2020}
\usepackage{times}
\usepackage{latexsym}

\usepackage{url}

\usepackage{tikz-dependency}
\usepackage{enumitem}
\usepackage{bm}
\usepackage{amsmath}
\usepackage{amssymb}
\usepackage{graphicx}
\graphicspath{ {./plots/} }
\usepackage{subfig}
\usepackage{color}
\usepackage{tabularx}
\usepackage{multirow}
\usepackage{xfrac}

\usepackage{microtype}

\aclfinalcopy 
\def\aclpaperid{1542} 

\definecolor{DarkRed}{RGB}{140,5,0}

\newcommand{\knn}{$k$-NN}

\newcommand{\ie}{i.\,e.\ }
\newcommand{\eg}{e.\,g.\ }

\title{Treebank Embedding Vectors for Out-of-Domain Dependency Parsing}

\author{Joachim Wagner \and James Barry \and Jennifer Foster \\
  ADAPT Centre \\ 
  School of Computing,
  Dublin City University, Ireland \\
  {\tt firstname.lastname@adaptcentre.ie}
}

\date{}

\begin{document}
\maketitle
\begin{abstract}

A recent advance in monolingual dependency parsing is the idea of a treebank embedding vector,
which allows all treebanks for a particular language to be used as training data while at the same time allowing the model to prefer training data from one treebank over others and
to select the preferred treebank at test time.
We build on this idea by 1) introducing a method to predict a treebank vector for sentences that do not come from a treebank used in training, and 2) exploring what happens when we move away from predefined treebank embedding vectors during test time and instead devise tailored interpolations.
We show that 1) there are interpolated vectors that are superior to the predefined ones, and
2) treebank vectors can be predicted with sufficient accuracy, for nine out of ten test languages,
to match  the performance of an oracle approach that knows the most suitable predefined treebank embedding for the test set.

\end{abstract}

\section{Introduction}
\label{sec:intro}

The Universal Dependencies project
\cite{nivre2016universal} has made available multiple treebanks for the same language annotated according to the same scheme, leading to a new wave of research which explores ways to use multiple treebanks in monolingual parsing
\cite{K17:3003,K17:3007,K17:3005,P18:2098}.

\newcite{P18:2098} introduced  a \textit{treebank embedding}. 
A single model is trained on the concatenation of the available treebanks for a language, and the input vector for each training token includes the treebank embedding which encodes the treebank the token comes from.  At test time, all input vectors in the test set of the same treebank are also assigned this treebank embedding vector.
\newcite{P18:2098} show that this approach is superior to mono-treebank training and to plain treebank concatenation. Treebank embeddings perform at about the same level as training on multiple treebanks and tuning on one, but they argue that a treebank embedding approach is preferable since it results in just one model per language.

What happens, however, when
the input sentence does not come from a treebank?  
\newcite{P18:2098} simulate this scenario with the Parallel Universal Dependency (PUD) test sets. 
They define the notion of a \textit{proxy} treebank  which is the treebank to be used for a treebank embedding when parsing sentences that do not come from any of the training treebanks. They empirically determine the best proxy treebank for each PUD test set by testing with each treebank embedding. 
However, the question remains what to do with sentences for which no gold parse is available, and for which we do not know the best proxy.

We investigate the problem of choosing treebank embedding vectors for new, possibly out-of-domain, sentences. In doing so, we explore the usefulness of \textit{interpolated} treebank vectors which are computed via a weighted combination of the predefined fixed ones. In experiments with Czech, English and French, 
we establish that useful interpolated treebank vectors exist. We then develop a simple k-NN method based on sentence similarity 
to choose a treebank vector, either fixed or interpolated, for sentences or entire test sets, which, for 9 of our 10 test languages matches the performance of the best (oracle) proxy treebank. 


\section{Interpolated Treebank Vectors}
\label{sec:tbemb}

Following recent work in neural dependency parsing
\cite{D14-1082,ballesteros-dyer-smith:2015:EMNLP,Q16-1023, 
zeman:EtAl:2017:K17:3, zeman:EtAl:2018:K18:2}, we represent an input token by concatenating various vectors.
In our experiments, each word $w_i$ in a sentence $S$ = $({w_1}$,...,${w_n})$ is a concatenation of 1) a dynamically learned word vector, 
2) a  word vector  obtained by passing the $k_i$ characters of ${w_i}$ through a BiLSTM 
and 3), following \newcite{P18:2098}, a treebank embedding
to distinguish the $m$ training treebanks:
\setlength{\abovedisplayskip}{4pt}
\setlength{\belowdisplayskip}{4pt}
\begin{equation}\label{tokvec}
\begin{aligned}
\mathbf{e}{(i)} ={} & \mathbf{e_1}{(w_i)} \\
 & \circ \mathbf{biLSTM}( \mathbf{e_2}{(ch_{i,1})}, ..., \mathbf{e_2}{(ch_{i,k_i})} ) \\
 & \circ \mathbf{f}
\end{aligned}
\end{equation}
\newcite{P18:2098} use 
\begin{equation}
   \mathbf{f} = e_3(t^\star) 
\end{equation}
where $t^\star \in {1, ..., m}$ is the source treebank for sentence $S$
or if $S$ does not
come from one of the $m$ treebanks, a choice of one of these (the proxy treebank). 
We change $\mathbf{f}$ during test time to
\begin{equation}
\mathbf{f} = \sum_{t=1}^{m} \alpha_t e_3(t)\label{dytbvec}
\end{equation}
where there are $m$ treebanks for the language in question and 
$\sum_{t=1}^{m}\alpha_t=1$.


\section{Data and Resources}
\label{sec:data}

For all experiments, we use UD v2.3 \cite{11234/1-2895}.
We choose Czech, English and French as our development languages because  they each have four treebanks (excluding  PUD), allowing us to train on three treebanks and test on a fourth. For testing, we use the PUD test sets for languages for which there are at least two other treebanks with training data: Czech, English, Finnish, French, Italian, Korean, Portuguese, Russian, Spanish and Swedish. 
Following \newcite{P18:2098}, we use the transition-based parser of
\newcite{delhoneux-stymne-nivre:2017:IWPT}
with the token input representations
as Eq.~\ref{tokvec} above.
Source code of our modified parser and helper scripts to carry out the experiments are available online.\footnote{\url{https://github.com/jowagner/tbev-prediction}}


\section{Are Interpolated Treebank Vectors Useful?}
\label{sec:useful}

We attempt to ascertain how useful interpolated treebank embedding vectors
are by 
examining the labelled attachment score (LAS) of trees parsed with different
interpolated treebank vectors. 
For each of our three development languages, 
we train multi-treebank parsing models
on the four combinations of three of the four available treebanks
and we test each model on the development sets of all four treebanks,
\ie three in-domain parsing settings and one out-of-domain setting.\footnote{An in-domain example
    is testing
    a model trained on
    \texttt{cs\_cac+cltt+fictree} on \texttt{cs\_cac}, and
    an out-of-domain example is
    testing the same model on \texttt{cs\_pdt}.
}

Since $m=3$ and $\sum_{t=1}^{m} \alpha_{t} = 1$, all treebank vectors lie in a plane and we can
visualise LAS results in colour plots.
As the treebank vectors can have arbitrary
distances, we plot (and sample) in the
weight space 
$\mathbf{R}^m$.
We include the equilateral triangle spanned by the three fixed treebank embedding vectors in our plots.
Points outside the triangle can be reached by
allowing negative weights $\alpha_t<0$.

We obtain treebank LAS and sentence-level LAS for 200 weight vectors sampled from the weight space,
including the corners of the triangle,
and repeat
with different seeds for parameter initialisation and training data shuffling.
Rather than sampling at random, points are chosen so that they are somewhat symmetrical and evenly distributed.

\begin{figure}
\begin{center}
\includegraphics[scale=0.50450]{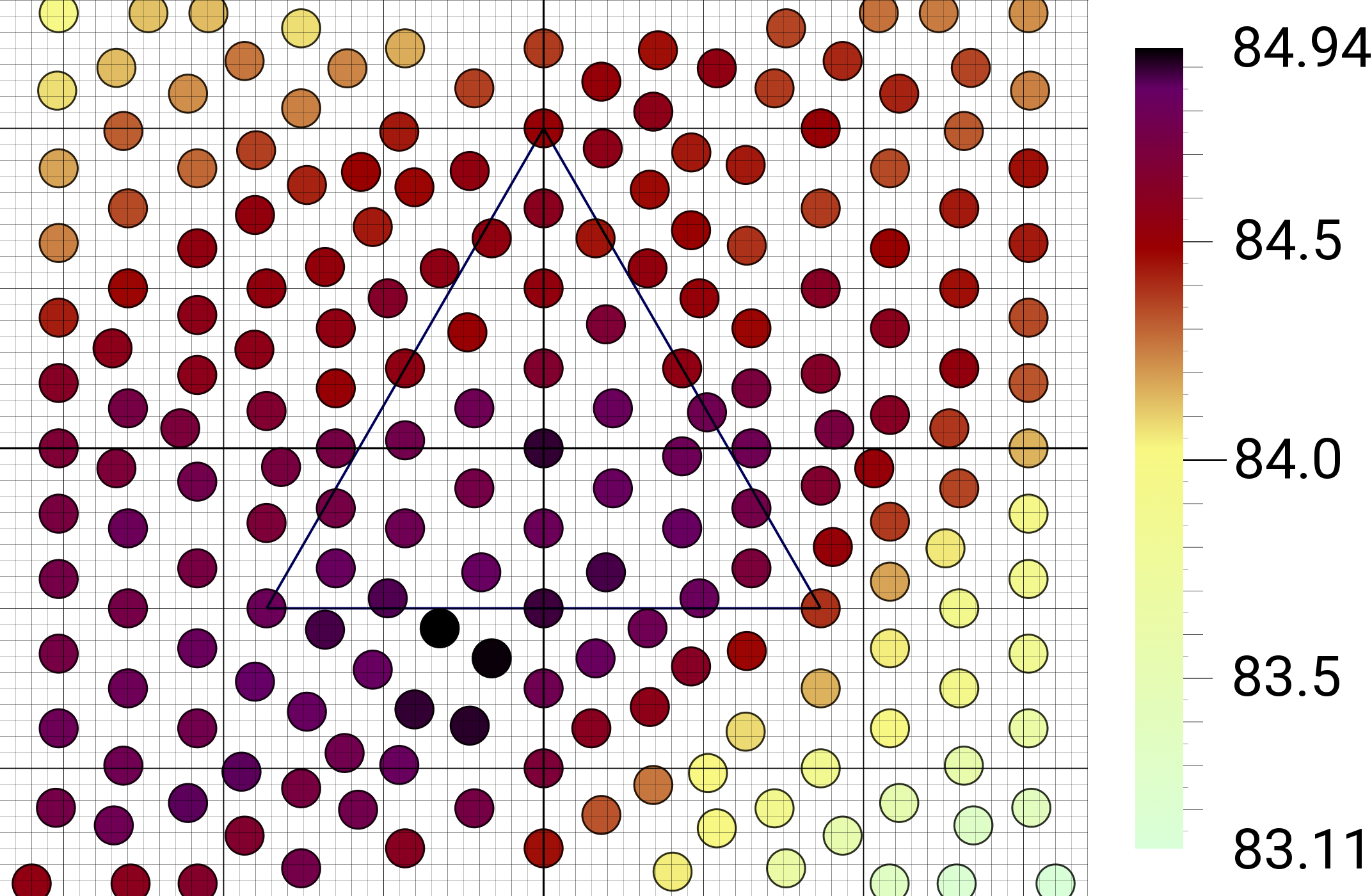}
\end{center}
\caption{LAS
in the treebank vector weight space ($m=3$) for
\texttt{cs\_cltt+fictree+pdt} on \texttt{cs\_cac-dev}
with the second seed.
}
\label{fig:useful:seeds}
\end{figure}

Figure~\ref{fig:useful:seeds} shows the development set LAS on
\texttt{cs\_cac-dev} for a model trained on
\texttt{cs\_cltt+fictree+pdt} with the second seed.
We create 432 such plots for nine seeds, four training configurations, four development sets 
and three languages.
The patterns vary with each seed and configuration.
The smallest LAS range within a plot is 87.8 to 88.3
(\texttt{cs\_cac+cltt+pdt} on \texttt{cs\_pdt} with the seventh seed).
The biggest LAS range is 59.7 to 76.8
(\texttt{fr\_gsd+sequoia+spoken} on \texttt{fr\_spoken} with the fifth seed).

The location of the fixed treebank vectors $e_3(t)$ are at the corners of the triangle in each graph.
For in-domain settings 
one or two corners usually have LAS close to the highest LAS in the plot.
The best LAS scores (black circles), however, are often
located
outside the triangle, \ie negative weights are needed to reach
it.

\begin{figure}
\begin{center}
\includegraphics[scale=0.50450]{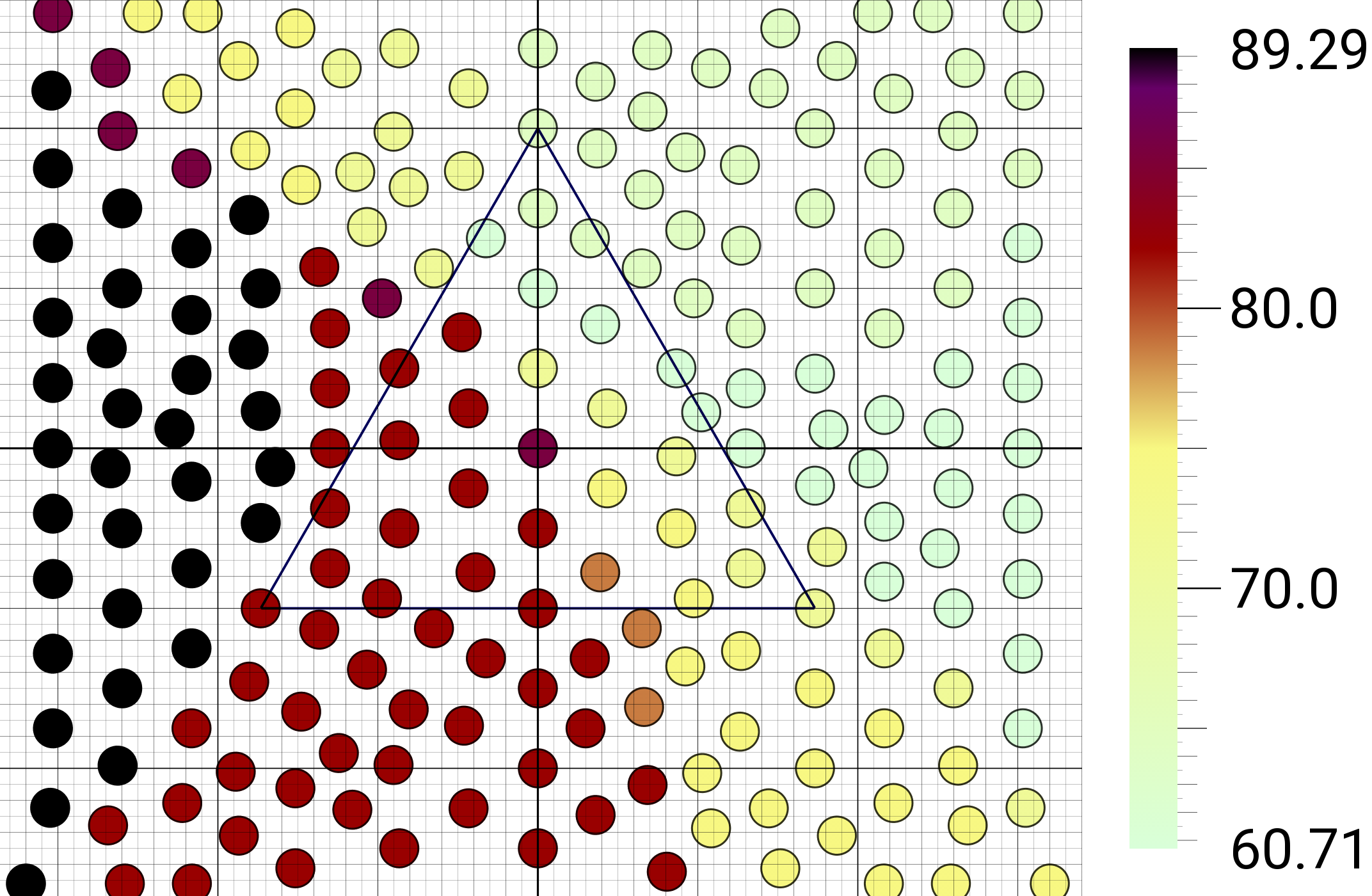}
\end{center}
\caption{LAS in the treebank vector weight space ($m=3$)
    for sentence 2 of \texttt{en\_partut-dev} (28 tokens) with 
    \texttt{en\_ewt+gum+lines} and our first seed.
}
\label{fig:useful:sentences}
\end{figure}

Turning to sentence-level LAS,
Figure~\ref{fig:useful:sentences} shows the LAS for an individual example sentence rather than an entire development set.
This sentence is taken from \texttt{en\_partut-dev} and
is parsed with a model trained on \texttt{en\_ewt+gum+lines}.
For this 28-token sentence, LAS can only change in steps of 1/28
and 34 of the 200 treebank embedding weight points
share the top score.
Negative weights are needed to reach these points
outside the triangle.

Over all development sentences and parsing models,
an interpolated treebank vector achieves highest LAS for 99.99\% of sentences:
In 78.07\% of cases, one of the corner vectors also achieves the highest LAS and in the remaining 21.92\%, interpolated vectors are needed.
It is also worth noting that, for 39\% of
sentences, LAS does not depend on the treebank vectors
at all, at least not in the weight range explored.

Often, LAS changes from one side to another side of the graph.
The borders have different orientation and sharpness.
The fraction of points with highest LAS varies from few to many.
The same is true for the fraction of points with lowest LAS.
Noise seems to be low.
Most data points match the performance of their neighbours, \ie
the scores are not sensitive to small changes of the treebank
weights, suggesting that the observed differences are not just
random numerical effects.

This preliminary analysis suggests that useful interpolated treebank vectors
do
exist. Our next step is to try to predict them.
In all subsequent experiments, we focus on the out-of-domain setting, \ie each multi-treebank model is tested on
a treebank not included in training.


\section{Predicting Treebank Vectors}
\label{sec:prediction}

We use $k$-nearest neighbour (\knn{}) classification to predict treebank embedding vectors for an individual sentence or a set of sentences at test time. We experiment with 1) allocating the treebank vector for an input \textit{sentence} using the $k$ most similar training \textit{sentences} (\texttt{se-se}), and 2) allocating the treebank vector for a \textit{set of input sentences} using the most similar training \textit{treebank} (\texttt{tr-tr}).

We will first explain the \texttt{se-se} case. For each input sentence, we retrieve from the
training data
the $k$ most similar sentences and then identify the treebank vectors from the candidate samples that have the highest LAS.
To compute similarity, we represent sentences either as tf-idf vectors computed over character 
n-grams, or as vectors produced by max-pooling
over a sentence's ELMo vectors~\cite{peters-EtAl:2018:N18-1}
produced by averaging all ELMo biLM layers.\footnote{We use
\textit{ELMoForManyLangs}~\cite{che-etal-2018-towards}.} 

We experiment with $k=1,3,9$. For many sentences, several treebank vectors yield the optimal LAS for the most similar retrieved sentence(s), and so we try several tie-breaking strategies, including choosing the vector closest to the uniform weight vector (\ie each of the three treebanks is equally weighted), re-ranking the list of vectors in the tie
according to the
LAS of the next most similar sentence, and using the average LAS of the $k$ sentences retrieved to 
choose
the treebank vector.
Three
treebank vector sample sizes were tried:
\begin{enumerate}
\setlength\itemsep{-0.5em}
    \item \texttt{fixed}: Only the three fixed treebank vectors, \ie the corners of the triangle in Fig.~\ref{fig:useful:seeds}.
     \item $\alpha_t \geq 0$: Negative weights are not used in the interpolation,
           \ie only the 32 points inside or on the triangle in  Fig.~\ref{fig:useful:seeds}.
     \item \texttt{any}: All 200 weight
     points shown in Fig.~\ref{fig:useful:seeds}.   
\end{enumerate}

When retrieving treebanks 
(\texttt{tr-tr}), we use the average of the treebank's sentence representation vectors as the treebank representation and we
normalise the vectors to the unit sphere as otherwise the size of the treebank would dominate the location in vector space.

We include oracle versions of each k-NN model in our experiments.
The k-NN oracle method is different from the normal k-NN method in that the test data is added to the training data so that the test data itself will be retrieved.
This means that a k-NN oracle with $k=1$ knows exactly what treebank vector is best for each test item while a basic k-NN model has to predict the best vector based on the training data.
In the \texttt{tr-tr} setting, our k-NN classifier is selecting
one of three treebanks for the fourth test treebank.
In the oracle k-NN setting, it selects the test treebank itself and parses the sentences in that treebank with its best-performing treebank vector.
When the treebank vector sample space is limited to the vectors for the three training treebanks (fixed), this method is the same as the best-proxy method of \newcite{P18:2098}.


\section{Results}

\begin{table}
\centering
\begin{tabular}{|ll|rrr|}
\multicolumn{2}{|c|}{\textbf{Model  (\texttt{se-se})}} & 
\multicolumn{3}{c|}{\textbf{Lang Avg LAS}} \\
\textbf{Learning} & \textbf{Weights}
 & \textbf{Cs} & \textbf{En} & \textbf{Fr}\\
\hline
random & fixed  & 82.5 & 73.4 & 72.1\\
random & $\alpha_t \geq 0$  & \textbf{82.6} & 73.9 & 72.5\\
random & any  & 82.4 & 73.3 & 72.1\\
\hline
k-NN & fixed    & 82.6 &         74.6  & \textbf{73.8} \\ 
k-NN & $\alpha_t \geq 0$  & 82.6 & \textbf{74.7} & 73.8\\
k-NN & any  & 82.6 & 74.4 & 73.7\\
\hline
oracle k-NN & fixed    & 84.1 & 77.8 & 77.1\\
oracle k-NN & $\alpha_t \geq 0$  & 84.2 & 79.3 & 78.6\\
oracle k-NN & any  & 85.5 & 81.0 & 80.2\\
\hline
\end{tabular}
\caption{Development set LAS  with per sentence treebank vectors
}
\label{devresults:avg:2}
\end{table}

The development results, averaged over the four development sets for each language, are shown in Tables~\ref{devresults:avg:2} and \ref{devresults:avg:1}.\footnote{To reduce noise from random initialisation, we parse each development set nine times with nine different seeds and use the median LAS.} As discussed above, upper bounds for \knn{} prediction are calculated by including an oracle setting in which the query item is added to the set of items to be retrieved, and $k$ restricted to 1. We are also curious to see what happens when an equal combination of the three fixed vectors (uniform weight vector) is used (\texttt{equal}),
and when treebank vectors are selected at random.

Table~\ref{devresults:avg:2} shows the \texttt{se-se} results.
The top section shows the results of randomly selecting a sentence's treebank vector, the middle section shows the \knn{} results and the bottom section the oracle \knn{} results.
The \knn{} predictor clearly outperforms the random predictor for English and French, but not for Czech, suggesting that the treebank vector itself plays less of a role for Czech, perhaps due to high domain overlap between the treebanks.
The oracle \knn{} results indicate not only the substantial room for improvement for the predictor, but also the potential of interpolated vectors  since the results improve as the sample space is increased beyond the three fixed vectors.

\begin{table}
\centering
\begin{tabular}{|ll|rrr|}
\multicolumn{2}{|c|}{\textbf{Model (\texttt{tr-tr})}} & 
\multicolumn{3}{c|}{\textbf{Lang Avg LAS}} \\
\textbf{Learning} & \textbf{Weights}
 & \textbf{Cs} & \textbf{En} & \textbf{Fr}\\
\hline
proxy-best & fixed  & 82.7 & 74.7 & 73.8\\
proxy-worst & fixed  & 82.3 & 72.4 & 70.7\\
\hline
k-NN & fixed    & \textbf{82.7} & 74.6 & 73.8 \\ 
k-NN & $\alpha_t \geq 0$  & 82.7 & 74.6 & \textbf{73.8}\\
k-NN & any  & 82.7 & 74.5 & 73.8\\
\hline
oracle k-NN & fixed  & 82.7 & 74.7 & 73.8\\ 
oracle k-NN &  $\alpha_t \geq 0$  & 82.8 & 75.1 & 74.2\\
oracle k-NN &  any  & 82.9 & 75.1 & 74.3\\
\hline
equal & n/a  & 82.7 & \textbf{74.8} & 72.9\\
\hline
\end{tabular}
\caption{Development set LAS with one treebank vector for all input sentences
}
\label{devresults:avg:1}
\end{table}

Table~\ref{devresults:avg:1} shows the \texttt{tr-tr} results.
The first section  is the proxy treebank embedding of \newcite{P18:2098} where one of the fixed treebank vectors is used for parsing the development set. We report the best- and worst-performing of the three (\texttt{proxy-best} and  \texttt{proxy-worst}). 
The $k$-NN methods are shown in the second section of Table~\ref{devresults:avg:1}.
The first row of this section (\texttt{fixed} weights) can be directly compared with the \texttt{proxy-best}.
For Czech and French, the \knn{} method matches the performance of
\texttt{proxy-best}.
For English, it comes close. Examining the per-treebank English results, \knn{} predicts the best proxy treebank for all but \texttt{en\_partut}, where it picks the second best (\texttt{en\_gum}) instead of the best (\texttt{en\_ewt}).

The oracle \knn{} results are shown in the third section of Table~\ref{devresults:avg:1}.\footnote{Recall
   that the first method in this section, \texttt{oracle fixed},
   is the same method as \texttt{proxy-best}.
}
Although less pronounced than for the more difficult \texttt{se-se} task, they indicate that there is still some room for improving the vector predictor at the document level if interpolated vectors are considered.

Our \texttt{equal} method, that
uses the weights (\sfrac{1}{3}, \sfrac{1}{3}, \sfrac{1}{3}),
is shown in the last row of Table~\ref{devresults:avg:1}. It 
is the overall best English model. Our best model for Czech is a \texttt{tr-tr} model which just selects from the three fixed treebank vectors. For French, the best is a \texttt{tr-tr} model which selects from interpolated vectors with positive weights. 
For the PUD languages not used in development, we  select the hyper-parameters based on average LAS on all 12 development sets. The resulting generic hyper-parameters
are the same as those for the best French model: \texttt{tr-tr} with interpolated vectors and positive weights.\footnote{While the \knn{} models selected for final testing use char-$n$-gram-based sentence representations, ELMo representations are competitive.}



The PUD test set results  are shown in Table~\ref{pudresults}. 
For nine out of ten languages we match
the oracle method \texttt{proxy-best}
within a 95\% confidence interval.\footnote{Statistical significance is tested with udapi-python (\url{https://github.com/udapi/udapi-python}).}
For Russian, the treebank vector of the second-best proxy treebank is chosen, falling 0.8 LAS points behind.
Still, this difference is not significant (p=0.055).
For English, the generic model also 
picks the second-best proxy treebank.\footnote{For
    Korean PUD, LAS scores are surprisingly low given that development results on \texttt{ko\_gsd} and \texttt{ko\_kaist} are above 76.5 for all seeds.
    A run with a mono-treebank model confirms low performance on Korean PUD. According to a reviewer, there are known differences
    in the annotation between the Korean UD treebanks.
}



\begin{table}
\centering
\begin{tabular}{l|r|rr|rr}
  &     & & &  & \textbf{lan-} \\
\textbf{lan-}  &     & \multicolumn{2}{c|}{\textbf{proxy}} &  \textbf{ge-} & \textbf{guage-} \\
\textbf{guage} & $m$ & \textbf{worst} & \textbf{best}      & \textbf{neric} & \textbf{specific} \\
\hline
cs &4 & 81.6 & \textbf{82.5} & \textbf{82.5} & \textbf{82.5} \\ 
en &4 & 76.4 & \textbf{82.9} & 80.7$^\dagger$ & 81.7$^\dagger$ \\ 
es &2 & 76.1 & \textbf{80.3} & \textbf{80.3} & -- \\ 
fi &2 & 52.5 & \textbf{80.6} & 80.5 & -- \\ 
fr &4 & 74.9 & \textbf{78.6} & \textbf{78.6} & \textbf{78.6} \\ 
it &3 & 84.4 & \textbf{85.5} & \textbf{85.5} & -- \\ 
ko &2 & 35.5 & 43.9 & \textbf{44.0} & -- \\ 
pt &2 & 74.6 & 77.4 & \textbf{77.6} & -- \\ 
ru &3 & 82.6 & \textbf{83.7} & 82.9 & -- \\ 
sv &2 & 73.7 & \textbf{74.7} & \textbf{74.7} & -- \\ 
\hline
\end{tabular}
\caption{PUD Test Set Results:
    Statistically significant differences between \texttt{proxy-best} and our best method are marked with 
    $^\dagger$
}
\label{pudresults}
\end{table}


\section{Conclusion}
\label{sec:conclusion}

In experiments with Czech, English and French, we investigated treebank embedding vectors, exploring the ideas of  interpolated vectors and vector weight prediction.
Our attempts to predict good vector weights using a simple regression model yielded encouraging results. Testing on PUD languages, we match the performance of using the best fixed treebank embedding vector in nine of ten cases within the bounds of statistical significance
and in five cases exactly match it. 

On the whole, it seems that our predictor is not yet good enough to find interpolated treebank vectors that are clearly superior to the basic, fixed vectors and that we know to exist from the oracle runs.
Still, we think it is encouraging that performance did not drop
substantially
when the set of candidate vectors was widened ($\alpha_t \geq 0$ and `any').
We do not think the superior treebank vectors found by the oracle runs are simply noise, \ie model fluctuations due to varied inputs, because the LAS landscape in the weight vector space is not noisy. For individual sentences, LAS is usually constant in large areas and there are clear, sharp steps to the next LAS level.
Therefore, we think
that there is room for improvement for the predictor to find interpolated vectors which are better than the fixed ones.
We plan to explore 
other methods to predict treebank vectors, \eg  neural sequence modelling, and to 
apply our ideas to the related task of language embedding prediction for zero-shot learning.

Another area for future work is to explore
what information treebank vectors encode.
The previous work on the use of treebank vectors in mono- and multi-lingual parsing suggests that treebank vectors encode information that enables the parser to select treebank-specific information where needed while also taking advantage of treebank-independent information available in the training data.
The type of information will depend on the selection of treebanks,
\eg in a polyglot setting the vector may simply encode the language, and in a monolingual setting such as ours it may encode annotation or domain differences between the treebanks.

Interpolating treebank vectors adds a layer of opacity, and, in future work, it would be interesting to carry out experiments with synthetic data, \eg varying the number of unknown words, to get a better understanding of what they may be capturing.

Future work should also test even simpler strategies which do not use the
LAS of previous parses to gauge the best treebank vector, \eg
always picking the largest treebank.


\section*{Acknowledgments}

This  research  is  supported  by  Science  Foundation Ireland
through the ADAPT Centre for Digital Content Technology, which is
funded under the SFI Research Centres Programme (Grant 13/RC/2106)
and is co-funded under the European Regional Development Fund.
We thank the
reviewers
for their inspiring questions and
detailed feedback.


\bibliographystyle{acl_natbib}
\bibliography{main}

\end{document}